\documentclass[10pt, conference]{IEEEtran}
\IEEEoverridecommandlockouts
\usepackage{cite}
\usepackage{amsmath,amssymb,amsfonts}
\usepackage{algorithmic}
\usepackage{algorithm}
\usepackage{enumitem}
\usepackage{graphicx}
\usepackage{textcomp}
\usepackage{xcolor}
\usepackage{booktabs}
\def\BibTeX{{\rm B\kern-.05em{\sc i\kern-.025em b}\kern-.08em
    T\kern-.1667em\lower.7ex\hbox{E}\kern-.125emX}}
\renewcommand{\IEEEauthorrefmark}[1]{\ensuremath{^{#1}}}
    
\begin{document}

\title{FairGC: Fairness-aware Graph Condensation}

\author{\IEEEauthorblockN{Yihan Gao\IEEEauthorrefmark{1},
Chenxi Huang\IEEEauthorrefmark{1}, Wen Shi\IEEEauthorrefmark{1}, Ke Sun\IEEEauthorrefmark{2}, Ziqi Xu\IEEEauthorrefmark{3}, Xikun Zhang\IEEEauthorrefmark{3}, Mingliang Hou\IEEEauthorrefmark{4}, Renqiang Luo\IEEEauthorrefmark{1}\textsuperscript{*}}
\IEEEauthorblockA{\IEEEauthorrefmark{1}College of Computer Science and Technology, Jilin University, Changchun, China}
\IEEEauthorblockA{\IEEEauthorrefmark{2}Key Laboratory of Advanced Design and Intelligent Computing, Dalian University, Dalian, China}
\IEEEauthorblockA{\IEEEauthorrefmark{3}School of Computing Technologies, RMIT University, Melbourne, Australia}
\IEEEauthorblockA{\IEEEauthorrefmark{4}Guangdong Institute of Smart Education, Jinan University, Guangzhou, China}
\IEEEauthorblockA{Email: \{gaoyh5523, huangcx5523, shiwen24\}@mails.jlu.edu.cn, sunke@dlu.edu.cn \\ \{ziqi.xu, xikun.zhang\}@rmit.edu.au, teemohold@outlook.com, lrenqiang@jlu.edu.cn}
\thanks{\textsuperscript{*}Corresponding author}
}


\maketitle

\begin{abstract}
Graph condensation (GC) has become a vital strategy for scaling Graph Neural Networks by compressing massive datasets into small, synthetic node sets. While current GC methods effectively maintain predictive accuracy, they are primarily designed for utility and often ignore fairness constraints. Because these techniques are bias-blind, they frequently capture and even amplify demographic disparities found in the original data. This leads to synthetic proxies that are unsuitable for sensitive applications like credit scoring or social recommendations. To solve this problem, we introduce FairGC, a unified framework that embeds fairness directly into the graph distillation process. Our approach consists of three key components. First, a Distribution-Preserving Condensation module synchronizes the joint distributions of labels and sensitive attributes to stop bias from spreading. Second, a Spectral Encoding module uses Laplacian eigen-decomposition to preserve essential global structural patterns. Finally, a Fairness-Enhanced Neural Architecture employs multi-domain fusion and a label-smoothing curriculum to produce equitable predictions. Rigorous evaluations on four real-world datasets, show that FairGC provides a superior balance between accuracy and fairness. Our results confirm that FairGC significantly reduces disparity in Statistical Parity and Equal Opportunity compared to existing state-of-the-art condensation models. The codes are available at https://github.com/LuoRenqiang/FairGC.
\end{abstract}


\section{Introduction}
\par Graph learning has become a fundamental framework for modeling complex interactions across various domains~\cite{xia2025graph,du2025telling}, yet the rapid expansion of modern datasets has made training standard Graph Neural Networks (GNNs) increasingly difficult due to heavy memory and time costs~\cite{luo2025fairgp}.
To overcome these computational barriers, graph condensation (GC) has emerged as a powerful paradigm that compresses massive, large-scale graphs into a tiny set of synthetic nodes while ensuring that models trained on this compact proxy achieve performance comparable to the original data~\cite{gao2025robgc}.
This technique is currently finding widespread use in data-heavy applications, such as analyzing billion-scale social networks to map user connections or processing massive financial transaction graphs for real-time risk assessment~\cite{zhang2025effective}.

\par Despite the technical maturity of mainstream GC paradigms, these frameworks are primarily engineered to maximize predictive utility while remaining largely indifferent to the underlying fairness constraints~\cite{luo2024algorithmic}.
Since these methods operate in a bias-blind manner, they often inadvertently capture and even amplify demographic disparities present in the original data, leading to synthetic surrogates that are unsuitable for socially sensitive applications~\cite{dong2023fairness,xu2022assessing,oldfield2025revisiting}.
In practical tasks like credit risk assessment or social friendship recommendation, this lack of fairness awareness results in significant gaps in Statistical Parity ($\Delta_\text{SP}$)~\cite{dwork2012fairness} and Equal Opportunity ($\Delta_\text{EO}$)~\cite{hardt2016equality}, where protected groups may suffer from lower positive prediction rates or disproportionately higher misclassification errors.
Ultimately, the failure to decouple predictive signals from sensitive attributes during the condensation process yields a biased proxy that fails to provide equitable decisions, highlighting a critical research gap in developing fairness-aware data reduction techniques.

\begin{figure}[t]
    \centering
    \includegraphics[width=0.45\textwidth]{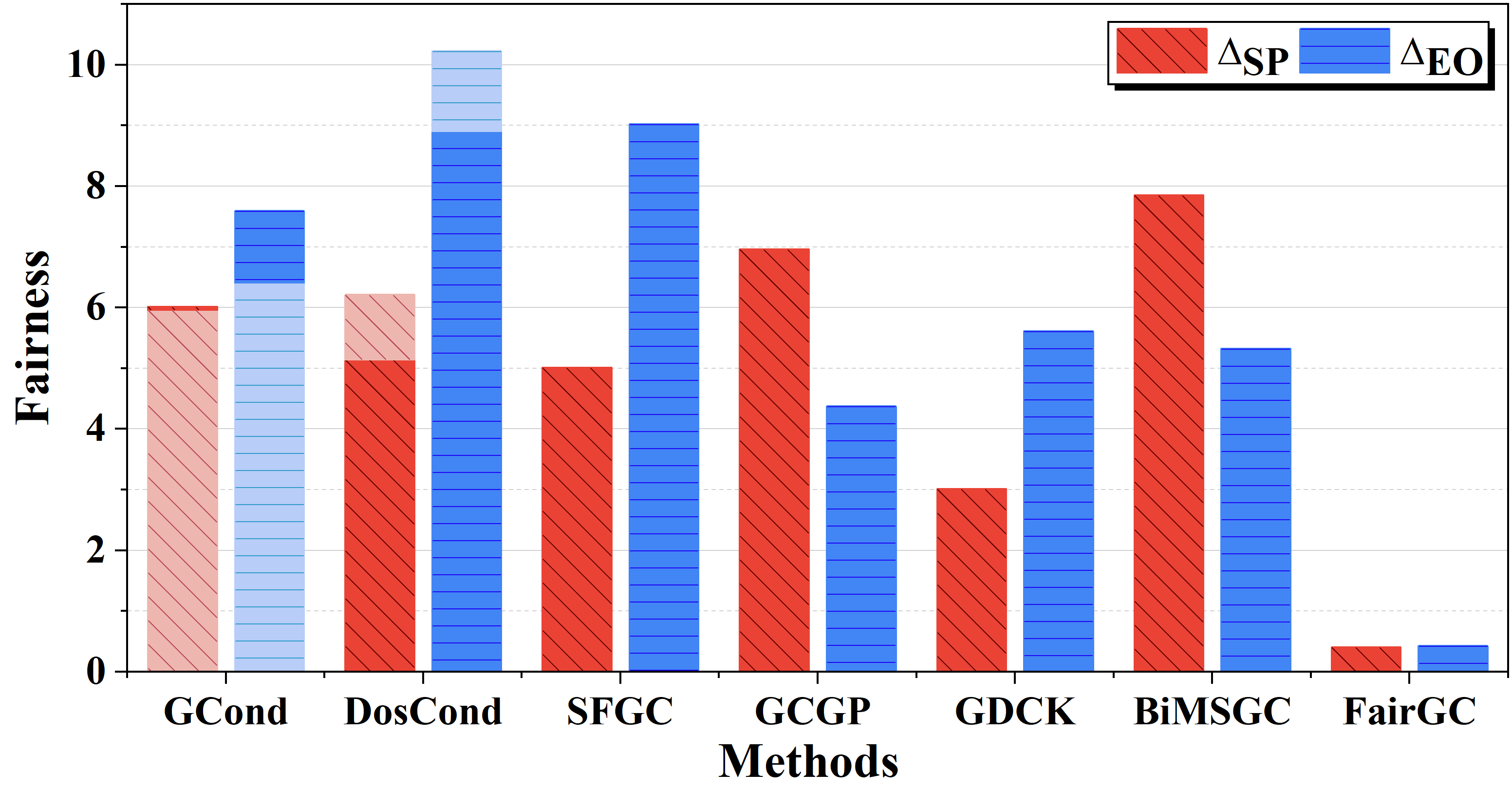}
    \caption{Fairness evaluation of GC methods on the Pokec-n-G dataset. 
    Both vanilla GC methods (darker bars) and their fairness-enhanced versions integrated with FairGNN (lighter bars) exhibit worse fairness compared to our proposed FairGC. 
    $\Delta_{\text{SP}}$ and $\Delta_{\text{EO}}$ are employed as fairness metrics (refer to Section~\ref{sec:FEM}), where lower values represent better fairness. 
    }
    \label{fig:back}
\end{figure}

\par To quantitatively demonstrate the fairness risks inherent in existing GC techniques, we evaluated six GC methods (GCond~\cite{jin2022graph}, DosCond~\cite{jin2022condensing}, SFGC~\cite{zheng2023structure}, GCGP~\cite{wang2025efficient}, GDCK~\cite{zhang2025gdck}, BiMSGC~\cite{fu2025bi}) on the Pokec-n-G dataset. 
Our experimental setup followed established fair graph learning protocols, employing $\Delta_\text{SP}$ and $\Delta_\text{EO}$ to measure performance gaps across sensitive groups. 
The results reveal that even when these frameworks are integrated with FairGNN~\cite{dai2023learning}, a widely used fairness-enhancing backbone, they still exhibit significantly higher disparities compared to our proposed FairGC.
This empirical evidence underscores a critical flaw: current methods treat data reduction and fairness as decoupled objectives, resulting in synthetic graphs that are fundamentally unsuitable for socially sensitive applications.

\par Crucially, existing GC paradigms fail to fundamentally resolve fairness concerns, and even the integration of external fairness-aware mechanisms often results in a degradation of overall performance rather than an improvement~\cite{luo2025fairness}.
This occurs because standard condensation algorithms are designed to maximize utility by matching gradients or distributions, which inadvertently discards the subtle structural dependencies and feature correlations essential for fairness constraints.
Since the condensation process lacks the inherent capability to carry fairness properties through the data reduction phase, the resulting synthetic nodes become biased proxies that downstream fair learners cannot rectify.
These observations emphasize the urgent need for a unified framework that can integrate fairness as a core objective of the condensation process itself, ensuring that the compressed graph is both efficient and socially equitable.

\par To bridge this critical gap, we propose FairGC, a novel framework that first utilizes a Distribution-Preserving Condensation module to synchronize label and sensitive attribute distributions, effectively preventing bias propagation during the data reduction phase.
Subsequently, it incorporates a Spectral Encoding module that leverages Laplacian eigen-decomposition and sinusoidal mapping to capture multi-scale topological footprints within the synthetic representation.
These distilled spatial and spectral insights are then integrated through a Fairness-Enhanced Neural Architecture that employs multi-domain fusion and a label-smoothing curriculum to ensure equitable downstream predictions.
Extensive experimental evaluations on diverse real-world datasets demonstrate that FairGC consistently achieves a superior balance between predictive utility and demographic fairness, significantly outperforming existing competitive GC paradigms.
Our main contributions can be summarized as follows:
\begin{itemize} [leftmargin=0.5cm]
\item We conduct an empirical study revealing that current GC methods amplify demographic bias and ``crystallize" it within synthetic nodes, making it difficult for downstream models to fix. 
Our analysis proves that existing fairness-aware learners fail to rectify these disparities once the data reduction process is complete. 
\item We introduce FairGC, a unified framework that embeds fairness directly into the distillation pipeline through distribution-preserving condensation and spectral encoding. 
This design ensures synthetic graphs are both topologically accurate and socially equitable, balancing model utility with demographic fairness. 
\item Extensive evaluations on real-world datasets like Credit and Pokec demonstrate that FairGC outperforms state-of-the-art baselines across various sensitive attributes. 
The results confirm our model achieves a superior trade-off between classification accuracy and fairness metrics for responsible graph learning. 
\end{itemize}

\section{Related Work}

\subsection{Graph Condensation}
\par GC emerges as a pivotal technique to distill large-scale graph structures into a significantly smaller set of synthetic nodes, ensuring that models trained on this compact proxy achieve comparable performance to those trained on the full dataset~\cite{zhang2022sparsified, zhang2025nearoptimal,zhang2023ricci}.
GDCK~\cite{zhang2025gdck} accelerates graph distillation by leveraging Neural Tangent Kernels and kernel ridge regression, bypassing the need for expensive GNN training and nested optimization loops.
GCGP~\cite{wang2025efficient} leverages Gaussian Processes to estimate predictive posterior distributions, eliminating iterative GNN training and enabling efficient gradient-based optimization of discrete graph structures via Concrete relaxation.
Traditional frameworks often maximize accuracy at the cost of exacerbating source graph biases, making their synthetic surrogates unsuitable for sensitive tasks. 
FairGC addresses this by coupling fairness with graph condensation.

\subsection{Fairness-aware GNNs}
\par Mitigating algorithmic bias in GNNs has become a crucial research frontier to prevent discriminatory protected groups~\cite{xu2025fairness}.
Pre-processing techniques~\cite{zhang2024endowing, zhu2024devil}, aim to rectify data-level bias by generating de-biased graph augmentations. 
In contrast, in-processing methods integrate fairness constraints directly into the training objective.
A seminal FairGNN, which leverages an adversarial debiasing framework to learn node representations that are invariant to sensitive attributes, even when such attributes are only partially observed. 
While other recent variants like FairSIN~\cite{yang2024fairsin} and FMP~\cite{jiang2024chasing} introduce sophisticated message-passing constraints, they are typically designed for static, full-scale graphs. 
These models often lack robustness for structural reduction, revealing a gap where fairness and data reduction are treated as isolated tasks.
FairGC fills this void by intrinsically coupling graph compression with fairness preservation.

\section{Preliminaries}

\subsection{Notations}
\par Let \( \mathcal{G} = (\mathcal{V}, \mathcal{E}, \mathbf{X}, \mathbf{y}, \mathbf{s}) \) denote the original graph, where \( \mathcal{V} \) is the set of \( n \) nodes, \( \mathcal{E} \) is the edge set, \( \mathbf{X} \in \mathbb{R}^{n \times d} \) is the node feature matrix, \( \mathbf{y} \in \{0,1\}^n \) is the label vector, and \( \mathbf{s} \in \{0,1\}^n \) is the binary sensitive attribute vector. 
The condensed graph is denoted as \( \mathcal{G}_{\mathrm{syn}} = (\mathcal{V}_{\mathrm{syn}}, \mathcal{E}_{\mathrm{syn}}, \tilde{\mathbf{X}}, \tilde{\mathbf{y}}, \tilde{\mathbf{s}}) \), where \( \tilde{\mathbf{X}} \in \mathbb{R}^{n_{\mathrm{syn}} \times d} \), \( \tilde{\mathbf{y}} \in \{0,1\}^{n_{\mathrm{syn}}} \), \( \tilde{\mathbf{s}} \in \{0,1\}^{n_{\mathrm{syn}}} \), and \( \mathcal{E}_{\mathrm{syn}} \) is represented by the adjacency matrix \( \mathbf{A}_{\mathrm{syn}} \in \{0,1\}^{n_{\mathrm{syn}} \times n_{\mathrm{syn}}} \).

\begin{figure*}[htbp]
	\centering
    \includegraphics[width=0.9\textwidth]{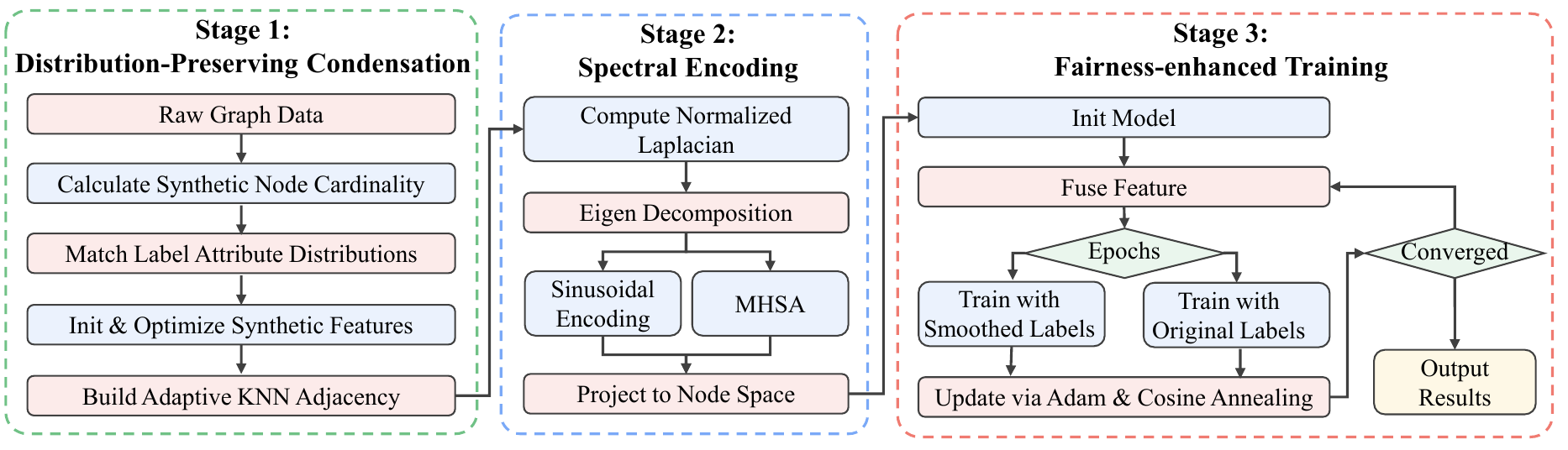}
    \caption{The illustration of FairGC.}
    \label{fig:architecture}
    \vspace{-1em}
\end{figure*}

\subsection{Fairness Evaluation Metrics} \label{sec:FEM}
\par To quantitatively evaluate the fairness of our proposed \text{FairGC}, we focus on two widely recognized independence-based criteria: Statistical Parity and Equal Opportunity.
In our notation, $y \in \{0,1\}$ and $\hat{y} \in \{0,1\}$ represent the ground-truth and predicted labels, respectively, while $s \in \{0,1\}$ indicates the sensitive attribute (e.g., gender or age).

\par Statistical Parity (SP) aims for an ideal state where the model's decision is decoupled from the sensitive group membership~\cite{dwork2012fairness}. 
A lower $\Delta_\text{SP}$ indicates that the model is more ``group-blind" in its global outcome distribution.
To measure the deviation from this equilibrium, we define it as follows:
\begin{equation}
\Delta_\text{SP} = \left| \mathbb{P}(\hat{y}=1 | s=0) - \mathbb{P}(\hat{y}=1 | s=1) \right|.
\label{equ:delta_SP}
\end{equation}

\par Equal Opportunity (EO) provides a more nuanced perspective by focusing on the fairness of qualified individuals~\cite{hardt2016equality}.
$\Delta_\text{EO}$ ensures that the model does not disproportionately misclassify positive instances from any specific protected group. 
Both metrics are computed based on the empirical probabilities observed in the test set.
Formally, the disparity in equal opportunity is captured by the difference in TPRs:
\begin{equation}
    \Delta_\text{EO}=|\mathbb{P}(\hat{y}=1|y=1,s=0)-\mathbb{P}(\hat{y}=1|y=1,s=1)|.
\label{equ:delta_EO}
\end{equation}

\subsection{Graph Condensation}
\par GC aims to distill a massive source graph $\mathcal{G}$ into a compact, synthetic counterpart $\mathcal{G}_{\mathrm{syn}}$ that retains equivalent training utility.
Distinct from graph sparsification or coreset selection, which are constrained to a subset of existing nodes, GC generates entirely new feature matrices and connectivity patterns.
This paradigm is typically formulated as an optimization problem intended to minimize the performance gap between models trained on original and synthetic data:
\begin{equation}
    \min_{\mathcal{G}_{\mathrm{syn}}} \mathcal{D} \left( \mathcal{M}(\mathcal{G}; \theta), \mathcal{M}(\mathcal{G}_{\mathrm{syn}}; \tilde{\theta}) \right),
\end{equation}
where $\mathcal{M}(\cdot)$ denotes a GNN-based learner, while $\theta$ and $\tilde{\theta}$ represent the parameters optimized on $\mathcal{G}$ and $\mathcal{G}_{\mathrm{syn}}$, respectively.

\section{Methods}
\subsection{Distribution-Preserving Condensation Module}
\par To alleviate computational overhead while upholding fairness constraints, we condense graph $\mathcal{G}=(\mathbf{X},\mathbf{y},\mathbf{s},\mathcal{E})$ into a synthetic surrogate $\mathcal{G}_{\text{syn}}=(\tilde{\mathbf{X}},\tilde{\mathbf{y}},\tilde{\mathbf{s}}, \mathbf{A}_{\text{syn}})$. 
Given a budget ratio $\rho \in (0,1)$, the synthetic cardinality is defined as:
\begin{equation}
n_{\text{syn}}=\max\left(10,\lfloor\rho \cdot n\rfloor\right).
\end{equation}

\subsubsection{Multi-faceted Distribution Synchronization}
\par To prevent bias propagation during condensation, we synchronize the marginal distributions of labels and sensitive attributes. 
Let $p_c$ and $q_a$ denote the empirical proportions of class $c$ and sensitive group $a$ in the training set. Node allocation is governed by:
\begin{equation}
n_{c,\text{syn}}=n_{\text{syn}} \times p_c, \quad n_{a,\text{syn}}=n_{\text{syn}} \times q_a.
\end{equation}

This ensures that $\tilde{\mathbf{y}}$ and $\tilde{\mathbf{s}}$ replicate the source statistical properties:
\begin{equation}
\mathbb{P}(\tilde{y}=c)=\mathbb{P}(y=c), \quad \mathbb{P}(\tilde{s}=a)=\mathbb{P}(s=a).
\end{equation}

For irregular sensitive attributes, randomized assignment is utilized to maintain distributional equilibrium.

\subsubsection{Feature Distillation via Proxy Optimization}
\par Synthetic features $\tilde{\mathbf{X}}$ are initialized via stratified sampling and stabilized through Z-score normalization:
\begin{equation}
\tilde{\mathbf{X}}^{(0)}=\frac{\mathbf{X}_{\text{sampled}}-\mu(\mathbf{X}_{\text{sampled}})}{\sigma(\mathbf{X}_{\text{sampled}})+\epsilon},
\end{equation}
where $\epsilon=10^{-8}$. We optimize $\tilde{\mathbf{X}}$ as learnable parameters by minimizing the empirical risk of a proxy MLP $g{\phi}$:
\begin{equation}
\mathcal{L}_{\text{cond}}=-\frac{1}{n_{\text{syn}}}\sum_{i=1}^{n_{\text{syn}}}\log[g_{\phi}(\tilde{\mathbf{x}}_i)]{\tilde{y}_i}.
\end{equation}

The optimization employs the Adam optimizer with gradient clipping to refine task-relevant utility.

\subsubsection{Hybrid Structural Reconstruction}
\par The adjacency matrix $\mathbf{A}_{\text{syn}}$ is reconstructed adaptively.
For large-scale settings ($n_{\text{syn}}>20,000$), we implement a sparse $k$-nearest neighbor (kNN) graph based on cosine similarity to ensure memory efficiency. 
Conversely, smaller graphs adopt a denser connectivity strategy.
This dual-track approach balances topological fidelity with scalability.

\begin{algorithm}[t]
\footnotesize
\caption{Fair Graph Condensation (FairGC)}
\label{alg:fairgc}
\begin{algorithmic}[1]
\REQUIRE Original graph $\mathcal{G}=(\mathbf{X}, \mathbf{y}, \mathbf{s}, \mathcal{E})$, compression ratio $\rho$, spectral components $K$, fusion layers $L$
\ENSURE Compact surrogate $\mathcal{G}_{\text{syn}}$ and fairness-enhanced model $f_{\theta}$

\STATE \COMMENT{Phase $1$: Distribution-Preserving Condensation}
\STATE Calculate $n_{\text{syn}}=\max\left(10,\lfloor\rho \cdot n\rfloor\right)$;
\STATE Compute $p_c$, $q_a$ from $\mathbf{y}$, $\mathbf{s}$;
\STATE Allocate $\tilde{\mathbf{y}}$, $\tilde{\mathbf{s}}$ via $n_{c,\text{syn}}=n_{\text{syn}} \times p_c, \quad n_{a,\text{syn}}=n_{\text{syn}} \times q_a$;
\STATE Initialize $\tilde{\mathbf{X}}^{(0)}=\frac{\mathbf{X}_{\text{sampled}}-\mu(\mathbf{X}_{\text{sampled}})}{\sigma(\mathbf{X}_{\text{sampled}})+\epsilon}$, where $\epsilon=10^{-8}$;
\WHILE{not converged}
    \STATE Compute $\mathcal{L}_{\text{cond}}=-\frac{1}{n_{\text{syn}}}\sum_{i=1}^{n_{\text{syn}}}\log[g_{\phi}(\tilde{\mathbf{x}}_i)]_{\tilde{y}_i}$;
    \STATE Update $\tilde{\mathbf{X}}$ via Adam with gradient clipping;
\ENDWHILE
\STATE Construct $\mathbf{A}_{\text{syn}}$ via adaptive kNN (sparse if $n_{\text{syn}}>20,000$, else dense);

\STATE \COMMENT{Phase $2$: Spectral Encoding}
\STATE Compute $\mathbf{L}_{\text{syn}} = \mathbf{I} - \mathbf{D}^{-\frac{1}{2}}\mathbf{A}_{\text{syn}}\mathbf{D}^{-\frac{1}{2}}$;
\STATE Eigendecompose $\mathbf{L}_{\text{syn}}$ to get $\{\lambda_i, \mathbf{u}_i\}_{i=1}^K$;
\STATE Generate $\mathbf{E}^{(0)}$ via sinusoidal encoding of $\lambda_i$;
\STATE Refine $\mathbf{E} = \text{LN}\left(\mathbf{E}' + \text{FFN}(\text{LN}(\mathbf{E}^{(0)} + \text{MHSA}(\mathbf{E}^{(0)})))\right)$;
\STATE Project: $\mathbf{z}_v^{\text{spec}} = \sum_{i=1}^{K} \mathbf{U}_{vi} \mathbf{E}_i$;

\STATE \COMMENT{Phase $3$: Fairness-Enhanced Training}
\STATE Initialize $f_{\theta}$ with FULayers, $\epsilon = 0.1$;
\FOR{$t=1$ to $T$}
    \STATE Fuse $\mathbf{H}$ and $\mathbf{Z}^{\text{spec}}$;
    \STATE $\mathcal{L} = \begin{cases} 
        \text{Smoothed NLL} & \text{if } t \leq 40 \\ 
        \text{Standard NLL} & \text{otherwise}
    \end{cases}$
    \STATE \quad \textit{where } $\tilde{\mathbf{y}}_i = (1-\epsilon)\mathbf{y}_i + \frac{\epsilon}{C}$;
    \STATE Update $\theta$ via AdamW with cosine annealing;
    \STATE Monitor $\Delta_{\mathrm{SP}}$, $\Delta_{\mathrm{EO}}$;
\ENDFOR
\RETURN $\mathcal{G}_{\text{syn}} = (\tilde{\mathbf{X}}, \tilde{\mathbf{y}}, \tilde{\mathbf{s}}, \mathbf{A}_{\text{syn}})$, $f_{\theta}$.
\end{algorithmic}
\end{algorithm}

\subsection{Spectral Encoding Module}
\par To preserve multi-scale structural characteristics, we propose a spectral encoding mechanism that captures global topological patterns.
Given $\mathbf{A}_{\text{syn}}$, we derive the normalized Laplacian matrix:
\begin{equation}
\mathbf{L}_{\text{syn}} = \mathbf{I} - \mathbf{D}^{-\frac{1}{2}}\mathbf{A}_{\text{syn}}\mathbf{D}^{-\frac{1}{2}},
\end{equation}
where $\mathbf{D}$ and $\mathbf{I}$ represent the degree and identity matrices, respectively. Eigendecomposition is performed to extract the leading $K$ eigenvalues and eigenvectors:
\begin{equation}
\mathbf{L}_{\text{syn}} = \mathbf{U}\mathbf{\Lambda}\mathbf{U}^{\top},\quad \mathbf{\Lambda} = \text{diag}(\lambda_1, \ldots, \lambda_K).
\end{equation}
\par The hyperparameter $K$ is calibrated per dataset via preliminary investigations to balance information retention and computational tractability.

\subsubsection{Sinusoidal Eigenvalue Encoding}
\par To transform discrete spectral values into a continuous space, we adopt a sinusoidal encoding scheme:
\begin{equation}
\begin{aligned}
\text{PE}(\lambda_i, 2j)     &= \sin\left(\frac{\lambda_i}{10000^{2j/d}}\right), \\
\text{PE}(\lambda_i, 2j+1)  &= \cos\left(\frac{\lambda_i}{10000^{2j/d}}\right),
\end{aligned}
\end{equation}
where $j \in [0, \dots, d/2-1]$. 
This mapping encapsulates the frequency characteristics of the graph spectrum.

\subsubsection{Attention-driven Spectral Refinement}
\par To model frequency dependencies, eigenvalue encodings $\mathbf{E}^{(0)}$ are processed via a multi-head self-attention (MHSA) layer:
\begin{equation}
\mathbf{E} = \text{LN}\left(\mathbf{E}' + \text{FFN}(\text{LN}(\mathbf{E}^{(0)} + \text{MHSA}(\mathbf{E}^{(0)})))\right),
\end{equation}
where $\text{LN}$ and $\text{FFN}$ denote Layer Normalization and Feed-Forward Network, respectively.

\subsubsection{Spectral Feature Projection}
\par Refined embeddings are projected onto the node space via the eigenvector basis:
\begin{equation}
\mathbf{z}_v^{\text{spec}} = \sum_{i=1}^{K} \mathbf{U}_{vi} \mathbf{E}_i.
\end{equation}

This projection bridges macroscopic topology and microscopic node features, ensuring the latent space is structurally informative.

\subsection{Fairness-Enhanced Neural Architecture}
\par We design a specialized architecture centered on multi-domain fusion and fairness-aware regularization to leverage distilled spatial and spectral knowledge.

\subsubsection{Feature Integration and Fusion}
\par Initial node features are projected via a standardized encoder: $\mathbf{H}^{(0)} = \sigma(\text{BN}(\mathbf{W}_0\mathbf{X}))$.
To facilitate cross-domain interaction, we introduce Fairness-aware Unified Layers (FULayer):
\begin{equation}
\mathbf{H}^{(l)} = \text{Norm}(\text{Dropout}(\sigma(\mathbf{W}_1^{(l)}\mathbf{H}^{(l-1)} + \mathbf{W}_2^{(l)}\mathbf{Z}^{\text{spec}}))),
\end{equation}
for $l = 1, \dots, L$. This iterative additive fusion ensures the final representation $\mathbf{H}^{(L)}$ incorporates global structural priors while maintaining node-level signals.

\subsubsection{Optimization and Fairness Alignment}
To balance utility and demographic alignment, we implement a Label Smoothing curriculum ($\epsilon=0.1$) for the initial $40$ epochs:
\begin{equation}
\tilde{\mathbf{y}}_i = (1-\epsilon)\mathbf{y}_i + \frac{\epsilon}{C},
\end{equation}

This soft-target approach encourages the learning of generalized, non-discriminatory decision boundaries.
Subsequent training utilizes standard negative log-likelihood (NLL).
Final predictions $\hat{\mathbf{y}}$ are generated via $\hat{\mathbf{y}} = \text{Softmax}(\text{MLP}(\mathbf{H}^{(L)}))$.
We employ the AdamW optimizer with cosine annealing and monitor metrics to ensure demographic unbiasedness.

\begin{table*}[htbp]
    \centering
    \footnotesize
    \caption{Experimental Results on Different Datasets. $\uparrow$ denotes the larger, the better; $\downarrow$ denotes the opposite. The best results are \textcolor{red}{\textbf{red and bold-faced}}. The runner-ups are \textcolor{blue}{\underline{blue and underlined}}.
    OOM means out-of-memory.}
    \setlength{\tabcolsep}{3pt}
    \renewcommand{\arraystretch}{0.9}
    \begin{tabular}{lccccccccc}
    \toprule
    & \multicolumn{3}{c}{Pokec-n-G (r: 0.05)} & \multicolumn{3}{c}{Pokec-n-R (r: 0.05)} & \multicolumn{3}{c}{Pokec-z-G (r: 0.05)} \\ \midrule
    Methods & ACC(\%) $\uparrow$ & $\Delta_{\text{SP}}$(\%) $\downarrow$ & $\Delta_{\text{EO}}$(\%) $\downarrow$ & ACC(\%) $\uparrow$ & $\Delta_{\text{SP}}$(\%) $\downarrow$ & $\Delta_{\text{EO}}$(\%) $\downarrow$ & ACC(\%) $\uparrow$ & $\Delta_{\text{SP}}$(\%) $\downarrow$ & $\Delta_{\text{EO}}$(\%) $\downarrow$ \\ \midrule
    GCond (ICLR '22) & $69.36_{\pm 0.94}$ & $6.01_{\pm 2.39}$ & $7.59_{\pm 3.33}$ & $69.32_{\pm 0.81}$ & $3.81_{\pm 0.52}$ & $6.57_{\pm 1.01}$ & $63.21_{\pm 1.23}$ & $5.40_{\pm 1.05}$ & $5.90_{\pm 1.64}$ \\
    DosCond (KDD '22) & $68.83_{\pm 1.19}$ & $5.14_{\pm 0.97}$ & $8.91_{\pm 2.27}$ & $68.54_{\pm 1.62}$ & $6.14_{\pm 0.64}$ & $6.94_{\pm 2.60}$ & $64.30_{\pm 0.69}$ & $5.87_{\pm 1.29}$ & $5.60_{\pm 1.14}$ \\
    SFGC (NeurIPS '23) & \textcolor{blue}{\underline{$70.45_{\pm 1.24}$}} & $5.01_{\pm 0.54}$ & $9.02_{\pm 2.01}$ & $69.82_{\pm 1.58}$ & $3.61_{\pm 0.86}$ & $1.83_{\pm 0.67}$ & $63.46_{\pm 1.57}$ & \textcolor{blue}{\underline{$1.47_{\pm 0.24}$}} & \textcolor{blue}{\underline{$2.84_{\pm 1.46}$}} \\
    GCGP (ArXiv '25) & \textcolor{blue}{\underline{$70.45_{\pm 0.95}$}} & $6.96_{\pm 0.98}$ & \textcolor{blue}{\underline{$4.37_{\pm 0.43}$}} & \textcolor{blue}{\underline{$70.39_{\pm 2.03}$}} & $4.68_{\pm 0.08}$ & $3.28_{\pm 0.36}$ & $64.40_{\pm 0.49}$ & $10.86_{\pm 0.31}$ & $8.35_{\pm 0.25}$ \\
    GDCK (PAKDD '25) & $70.13_{\pm 0.54}$ & \textcolor{blue}{\underline{$3.01_{\pm 0.59}$}} & $5.61_{\pm 0.83}$ & $70.31_{\pm 0.68}$ & $2.52_{\pm 0.57}$ & \textcolor{blue}{\underline{$2.97_{\pm 0.86}$}} & \textcolor{blue}{\underline{$64.48_{\pm 0.34}$}} & $2.47_{\pm 0.13}$ & $6.30_{\pm 0.17}$ \\
    BiMSGC (AAAI '25) & $69.53_{\pm 1.00}$ & $7.85_{\pm 1.72}$ & $5.32_{\pm 2.01}$ & $69.56_{\pm 1.25}$ & \textcolor{blue}{\underline{$1.49_{\pm 0.40}$}} & $6.16_{\pm 2.10}$ & $63.48_{\pm 1.24}$ & $1.53_{\pm 0.14}$ & $4.86_{\pm 1.29}$ \\
    GCond$_\text{+FairGNN}$ & $69.85_{\pm 0.56}$ & $5.96_{\pm 1.89}$ & $6.41_{\pm 3.27}$ & $69.75_{\pm 0.47}$ & $4.73_{\pm 0.81}$ & $6.19_{\pm 2.07}$ & $63.13_{\pm 0.64}$ & $6.72_{\pm 0.77}$ & $7.58_{\pm 1.24}$ \\
    DosCond$_\text{+FairGNN}$ & $69.41_{\pm 0.18}$ & $6.21_{\pm 0.37}$ & $10.22_{\pm 2.15}$ & $69.15_{\pm 0.58}$ & $4.44_{\pm 1.51}$ & $3.26_{\pm 0.85}$ & $64.15_{\pm 0.67}$ & $7.73_{\pm 4.21}$ & $6.08_{\pm 5.11}$ \\ 
    FairGC & \textcolor{red}{$\mathbf{70.87_{\pm 0.28}}$} & \textcolor{red}{$\mathbf{0.40_{\pm 0.23}}$} & \textcolor{red}{$\mathbf{0.42_{\pm 0.11}}$} & \textcolor{red}{$\mathbf{70.47_{\pm 0.14}}$} & \textcolor{red}{$\mathbf{0.58_{\pm 0.08}}$} & \textcolor{red}{$\mathbf{0.84_{\pm 0.13}}$} & \textcolor{red}{$\mathbf{64.78_{\pm 0.28}}$} & \textcolor{red}{$\mathbf{0.82_{\pm 0.62}}$} & \textcolor{red}{$\mathbf{0.72_{\pm 0.51}}$} \\
    \midrule
    & \multicolumn{3}{c}{Pokec-z-R (r: 0.05)} & \multicolumn{3}{c}{Credit (r: 0.01)} & \multicolumn{3}{c}{AMiner-L (r: 0.05)} \\ \midrule
    Methods & ACC(\%) $\uparrow$ & $\Delta_{\text{SP}}$(\%) $\downarrow$ & $\Delta_{\text{EO}}$(\%) $\downarrow$ & ACC(\%) $\uparrow$ & $\Delta_{\text{SP}}$(\%) $\downarrow$ & $\Delta_{\text{EO}}$(\%) $\downarrow$ & ACC(\%) $\uparrow$ & $\Delta_{\text{SP}}$(\%) $\downarrow$ & $\Delta_{\text{EO}}$(\%) $\downarrow$ \\ 
    \midrule
    GCond (ICLR '22) & $62.51_{\pm 1.41}$ & $3.35_{\pm 1.35}$ & $6.65_{\pm 0.88}$ & $77.77_{\pm 0.05}$ & $0.59_{\pm 0.27}$ & $0.73_{\pm 0.28}$ & $89.29_{\pm 0.28}$ & $8.06_{\pm 3.63}$ & $9.28_{\pm 1.09}$ \\
    DosCond (KDD '22) & $64.34_{\pm 0.66}$ & $4.12_{\pm 1.61}$ & $4.94_{\pm 2.19}$ & $77.74_{\pm 0.05}$ & $0.96_{\pm 1.14}$ & $1.11_{\pm 1.25}$ & \textcolor{red}{$\mathbf{90.87_{\pm 0.24}}$} & $10.80_{\pm 1.12}$ & $8.46_{\pm 2.06}$ \\
    SFGC (NeurIPS '23) & \textcolor{blue}{\underline{$64.35_{\pm 0.86}$}} & $2.99_{\pm 1.53}$ & $10.58_{\pm 1.80}$ & $77.14_{\pm 1.26}$ & $6.14_{\pm 1.35}$ & $4.87_{\pm 1.32}$ & $90.30_{\pm 1.44}$ & $6.68_{\pm 1.32}$ & $8.35_{\pm 1.44}$ \\
    GCGP (ArXiv '25) & \textcolor{blue}{\underline{$64.35_{\pm 1.12}$}} & $10.86_{\pm 1.79}$ & $8.35_{\pm 1.24}$ & $77.82_{\pm 0.76}$ & $6.82_{\pm 1.65}$ & $3.92_{\pm 0.04}$ & OOM & OOM & OOM \\
    GDCK (PAKDD '25) & $64.07_{\pm 0.34}$ & \textcolor{blue}{\underline{$0.86_{\pm 0.53}$}} & \textcolor{blue}{\underline{$2.57_{\pm 0.77}$}} & $76.94_{\pm 2.16}$ & $6.03_{\pm 0.57}$ & $9.32_{\pm 1.60}$ & OOM & OOM & OOM \\
    BiMSGC (AAAI '25) & $63.74_{\pm 0.76}$ & $1.91_{\pm 0.02}$ & $3.48_{\pm 1.02}$ & \textcolor{red}{$\mathbf{78.13_{\pm 0.06}}$} & \textcolor{blue}{\underline{$0.56_{\pm 0.02}$}} & $1.61_{\pm 0.14}$ & $89.97_{\pm 1.43}$ & $8.58_{\pm 2.19}$ & $6.94_{\pm 2.15}$ \\
    GCond$_\text{+FairGNN}$ & $63.21_{\pm 0.80}$ & $3.46_{\pm 2.68}$ & $8.39_{\pm 3.12}$ & $77.06_{\pm 0.63}$ & $0.87_{\pm 0.92}$ & \textcolor{blue}{\underline{$1.12_{\pm 1.36}$}} & $86.21_{\pm 3.54}$ & $6.18_{\pm 6.98}$ & $9.40_{\pm 5.02}$ \\
    DosCond$_\text{+FairGNN}$ & $64.09_{\pm 0.35}$ & $3.88_{\pm 2.64}$ & $4.90_{\pm 2.76}$ & $77.04_{\pm 0.38}$ & $3.24_{\pm 2.93}$ & $3.50_{\pm 3.69}$ & $90.66_{\pm 0.09}$ & $9.48_{\pm 1.27}$ & $5.88_{\pm 2.31}$ \\ 
    FairGC & \textcolor{red}{$\mathbf{64.62_{\pm 0.30}}$} & \textcolor{red}{$\mathbf{0.10_{\pm 0.09}}$} & \textcolor{red}{$\mathbf{0.29_{\pm 0.28}}$} & \textcolor{blue}{\underline{$77.94_{\pm 0.10}$}} & \textcolor{red}{$\mathbf{0.20_{\pm 0.11}}$} & \textcolor{red}{$\mathbf{0.22_{\pm 0.06}}$} & \textcolor{blue}{\underline{$90.80_{\pm 0.33}$}} & \textcolor{red}{$\mathbf{1.06_{\pm 0.71}}$} & \textcolor{red}{$\mathbf{3.28_{\pm 1.26}}$} \\
    \bottomrule
    \end{tabular}
    \label{tab:updated_dataset_results}
\end{table*}

\section{Experiments}

\subsection{Datasets} 
\par We evaluate the performance of FairGC using four diverse real-world graphs: Credit \cite{agarwal2021towards}, Pokec-z, Pokec-n \cite{takac2012data}, and AMiner-L \cite{tang2008arnetminer}. 
Credit is a financial transaction network where edges indicate spending similarities between users; binarized Age is utilized as the sensitive attribute for default risk prediction. 
Pokec-z and Pokec-n are social friendship networks from Slovakian regions, Zilinsky and Nitriansky. 
Following common fairness benchmarks, we predict users' employment categories while considering Gender (Pokec-z-G and Pokec-n-G) and Region (Pokec-z-R and Pokec-n-R) as sensitive features.
AMiner-L is a large-scale co-authorship network where the task is to predict research fields, using Affiliation Continent as the protected attribute.
All experiments were conducted on a workstation equipped with a server equipped with $3$× NVIDIA L$40$ GPUs with the driver version $580.95.05$.

\subsection{Baselines}
\par In our experiments, we compare FairGC against a comprehensive set of SOTA baselines, which are categorized into two groups: 1) GC methods that focus on distilling large-scale graph structures into a significantly smaller set of synthetic nodes, and 2) fairness-aware GNNs that aim to mitigate bias in graph learning. 
For the fairness-aware GNNs, since they are not originally designed for GC, we adapt each model using the different GC framework, to enable a fair comparison under condensation scenarios, ensuring consistency in handling distilling requests.

\par We list six GC baselines: GCond~\cite{jin2022graph}, DosCond~\cite{jin2022condensing}, SFGC~\cite{zheng2023structure}, GCGP~\cite{wang2025efficient}, 
GDCK~\cite{zhang2025gdck},
BiMSGC~\cite{fu2025bi}.
We use one fairness-aware GNN baseline with GC adaptation (some of them could not suit for fairness-aware controlling): FairGNN~\cite{dai2023learning}.
These baselines represent the current SOTA in GC and fairness-aware graph learning. 
By adapting the FairGNN with GC capability, we ensure a comprehensive and fair evaluation of FairGC against methods that address both fairness and condensation challenges.

%

\subsection{Comparison Results}
\par In this section, we present a comprehensive evaluation of FairGC against state-of-the-art GC baselines across several real-world datasets: Pokec-n, Pokec-z, Credit, and AMiner-L. 
The comparison focuses on the balance between predictive accuracy (ACC) and fairness metrics ($\Delta_{\text{SP}}$ and $\Delta_{\text{EO}}$). 
The results, summarized in Table~\ref{tab:updated_dataset_results}, demonstrate that FairGC consistently achieves a superior trade-off between utility and fairness compared to all existing methods.

\par \textbf{Predictive Accuracy.} FairGC maintains highly competitive node classification accuracy across all datasets, frequently achieving the best or second-best performance. 
For instance, on Pokec-n-G and Pokec-z-G, FairGC achieves the highest accuracy ($70.87\%$ and $64.78\%$, respectively), outperforming advanced baselines like BiMSGC and GCond. 
On the Credit and AMiner-L datasets, while FairGC’s accuracy is marginally lower than the top-performing utility-centric methods (e.g., $90.80\%$ vs. $90.87\%$ for DosCond on AMiner-L), this minor difference is negligible. 
This indicates that our fairness-aware condensation process preserves the essential informative features of the original graph without significant degradation in predictive power.

\par \textbf{Fairness Performance.} 
FairGC significantly enhances fairness by dramatically reducing both statistical parity and equal opportunity disparities. 
Across all evaluated datasets, FairGC achieves the lowest $\Delta_{\text{SP}}$ and $\Delta_{\text{EO}}$ values, often by a substantial margin. 
For example, on Pokec-n-G, FairGC reduces $\Delta_{\text{EO}}$ to $0.42\%$, a nearly 18x improvement over the standard GCond ($7.59\%$).
Notably, even when baselines are augmented with fairness-aware backbones (e.g., GCond$_{\text{+FairGNN}}$), they still struggle to match FairGC's performance. 
On the Credit dataset, FairGC achieves a $\Delta_{\text{EO}}$ of $0.22\%$, while most baselines fluctuate between $0.5\%$ and $1.1\%$. 
These results underscore that FairGC effectively eliminates discriminatory patterns that traditional condensation methods often inadvertently preserve or amplify.

\par Collectively, FairGC emerges as a robust solution that excels in both accuracy and fairness. 
Unlike utility-focused methods (e.g., DosCond, SFGC) that achieve high accuracy at the cost of high bias, or simple fairness-augmented baselines that fail to achieve optimal equity, FairGC provides a holistic framework. 
It effectively addresses the challenges of preserving distribution-level fairness during graph reduction.
The consistent gains across diverse datasets and condensation ratios validate FairGC's scalability and its practical value for socially sensitive applications where ethical considerations are as critical as model performance.

\begin{table}[t]
    \centering
    \footnotesize
    \caption{Ablation study results for FairGC. ``P" and ``AM" denote the Pokec and AMiner datasets, respectively.}
    \label{tab:ablation_results}
    \setlength{\tabcolsep}{5pt}
    \renewcommand{\arraystretch}{0.9}
    \begin{tabular}{lcccc}
        \toprule
        Dataset & Metric & FairGC & FairGC w/o C & FairGC w/o F \\
        \midrule
        P-n-G & ACC (\%) & \textcolor{red}{$\mathbf{70.87_{\pm 0.28}}$} & $70.17_{\pm 0.05}$ & $69.36_{\pm 0.94}$ \\
        & $\Delta_{\text{SP}}$ (\%) & \textcolor{red}{$\mathbf{0.40_{\pm 0.23}}$} & $0.48_{\pm 0.26}$ & $6.01_{\pm 2.39}$ \\
        & $\Delta_{\text{EO}}$ (\%) & \textcolor{red}{$\mathbf{0.42_{\pm 0.11}}$} & $1.05_{\pm 0.85}$ & $7.59_{\pm 3.33}$ \\
        \midrule
        P-n-R & ACC (\%) & \textcolor{red}{$\mathbf{70.47_{\pm 0.14}}$} & $70.23_{\pm 1.06}$ & $69.32_{\pm 0.81}$ \\
        & $\Delta_{\text{SP}}$ (\%) & \textcolor{red}{$\mathbf{0.58_{\pm 0.08}}$} & $0.67_{\pm 0.25}$ & $3.81_{\pm 0.52}$ \\
        & $\Delta_{\text{EO}}$ (\%) & \textcolor{red}{$\mathbf{0.84_{\pm 0.13}}$} & $1.06_{\pm 0.66}$ & $6.57_{\pm 1.01}$ \\
        \midrule
        P-z-G & ACC (\%) & \textcolor{red}{$\mathbf{64.78_{\pm 0.28}}$} & $63.48_{\pm 0.97}$ & $63.21_{\pm 1.23}$ \\
        & $\Delta_{\text{SP}}$ (\%) & \textcolor{red}{$\mathbf{0.82_{\pm 0.62}}$} & $2.23_{\pm 1.18}$ & $5.90_{\pm 1.64}$ \\
        & $\Delta_{\text{EO}}$ (\%) & \textcolor{red}{$\mathbf{0.72_{\pm 0.51}}$} & $2.23_{\pm 1.18}$ & $5.90_{\pm 1.64}$ \\
        \midrule
        P-z-R & ACC (\%) & \textcolor{red}{$\mathbf{64.62_{\pm 0.30}}$} & $63.44_{\pm 1.02}$ & $62.51_{\pm 1.41}$ \\
        & $\Delta_{\text{SP}}$ (\%) & \textcolor{red}{$\mathbf{0.10_{\pm 0.09}}$} & $1.86_{\pm 1.03}$ & $3.35_{\pm 1.35}$ \\
        & $\Delta_{\text{EO}}$ (\%) & \textcolor{red}{$\mathbf{0.29_{\pm 0.28}}$} & $4.23_{\pm 2.17}$ & $6.65_{\pm 0.88}$ \\
        \midrule
        Credit & ACC (\%) & \textcolor{red}{$\mathbf{77.94_{\pm 0.10}}$} & $77.54_{\pm 0.14}$ & $77.77_{\pm 0.05}$ \\
        & $\Delta_{\text{SP}}$ (\%) & \textcolor{red}{$\mathbf{0.20_{\pm 0.11}}$} & $0.42_{\pm 0.33}$ & $0.59_{\pm 0.27}$ \\
        & $\Delta_{\text{EO}}$ (\%) & \textcolor{red}{$\mathbf{0.22_{\pm 0.06}}$} & $0.45_{\pm 0.30}$ & $0.73_{\pm 0.28}$ \\
        \midrule
        AM-L & ACC (\%) & \textcolor{red}{$\mathbf{90.80_{\pm 0.33}}$} & $88.71_{\pm 0.58}$ & $89.29_{\pm 0.11}$ \\
        & $\Delta_{\text{SP}}$ (\%) & \textcolor{red}{$\mathbf{1.06_{\pm 0.71}}$} & $1.56_{\pm 1.08}$ & $8.06_{\pm 3.63}$ \\
        & $\Delta_{\text{EO}}$ (\%) & \textcolor{red}{$\mathbf{3.28_{\pm 1.26}}$} & $4.08_{\pm 2.15}$ & $9.28_{\pm 1.09}$ \\
        \bottomrule
    \end{tabular}
\end{table}

\subsection{Ablation Study}
\par To evaluate FairGC’s components, we conduct an ablation study focusing on distribution-preserving condensation (FairGC w/o C) and fairness-enhanced architecture (FairGC w/o F), with results in Table~\ref{tab:ablation_results}. 
The FairGC w/o F variant consistently exhibits the poorest fairness, as $\Delta_{\text{SP}}$ and $\Delta_{\text{EO}}$ increase significantly across all datasets without explicit fairness-specific design. 
Similarly, the FairGC w/o C variant underscores the necessity of our condensation strategy; while traditional methods reduce graph size, they fail to preserve the nuanced statistical distributions required for fair downstream classification.

\par In summary, the ablation study underscores that both the distribution-preserving condensation module and the fairness-enhanced neural architecture are indispensable. 
While replacing the fairness architecture may occasionally boost accuracy, it leads to unacceptable fairness degradation. 
Likewise, traditional condensation fails to preserve the nuanced distributions necessary for fair representation.

\section{Conclusion}
\par In this paper, we address the critical research gap where existing GC methods prioritize predictive accuracy but inadvertently amplify demographic biases during data reduction. 
We propose FairGC, a novel framework that integrates fairness as a core objective by combining distribution-preserving condensation, spectral encoding for structural fidelity, and a fairness-enhanced neural architecture. 
Through extensive empirical analysis on real-world datasets like Credit and Pokec-n, we demonstrate that FairGC effectively decouples predictive signals from sensitive attributes, preventing the crystallization of bias in synthetic nodes. 
Our results confirm that FairGC achieves a superior balance between training utility and demographic equity, providing a robust and socially responsible solution for deploying graph intelligence in resource-constrained environments.

\section*{Acknowledgments}
This work was supported by the National Natural Science Foundation of China (NSFC) under Grant No.$62406054$.

\bibliographystyle{IEEEtran}
\bibliography{ref}
\end{document}